\pdfoutput=1

\documentclass[11pt]{article}

\usepackage[final]{acl}

\usepackage{times}
\usepackage{latexsym}

\usepackage[T1]{fontenc}

\usepackage[utf8]{inputenc}

\usepackage{microtype}
\usepackage{booktabs}
\usepackage{tabularx} 
\usepackage{amsmath}
\usepackage{multirow}
\usepackage{multicol}
\usepackage{graphics}
\usepackage{graphicx}
\usepackage{soul}
\usepackage{ulem}
\usepackage{color}
\usepackage{xcolor}
\newcommand{\hlred}{\colorlet{c}{red!20}\sethlcolor{c}\hl}

\title{Persona-Guided Planning for Controlling the Protagonist's Persona in Story Generation}

\author{
  Zhexin Zhang\footnotemark[1], 
  Jiaxin Wen\footnotemark[1], 
  Jian Guan, 
  Minlie Huang\footnotemark[2] \\
  The CoAI group, DCST; Institute for Artificial Intelligence; State Key Lab \\of Intelligent Technology and Systems;
Beijing National Research Center for\\ Information Science and Technology; 
Tsinghua University, Beijing 100084, China.\\
  \texttt{\{zx-zhang18, wenjx17, j-guan19\}@mails.tsinghua.edu.cn}\\
  \texttt{aihuang@tsinghua.edu.cn}}

\date{}

\begin{document}
\maketitle
\begin{abstract}

Endowing the protagonist with a specific personality is essential for writing an engaging story. In this paper, we aim to control the protagonist's persona in story generation, i.e., generating a story from a leading context and a persona description, where the protagonist should exhibit the specified personality through a coherent event sequence. Considering that personas are usually embodied implicitly and sparsely in stories, we propose a planning-based generation model named \textsc{ConPer} to explicitly model the relationship between personas and events. \textsc{ConPer} first plans events of the protagonist's behavior which are motivated by the specified persona through predicting one target sentence, then plans the plot as a sequence of keywords with the guidance of the predicted persona-related events and commonsense knowledge, and finally generates the whole story. Both automatic and manual evaluation results demonstrate that \textsc{ConPer} outperforms state-of-the-art baselines for generating more coherent and persona-controllable stories. Our code is available at \url{https://github.com/thu-coai/ConPer}.

\end{abstract}

\section{Introduction}

\renewcommand{\thefootnote}{\fnsymbol{footnote}}

\footnotetext[1]{Equal contribution.}
\footnotetext[2]{Corresponding author.}

\begin{table}[!t]
\scriptsize
\begin{tabular}{p{207pt}}
    \toprule
        \textsf{\textbf{Context:} $\cdots$ Artur Boruc, a Polish national pilot, was going to get the group infiltrated into the area and prepared for the attack. $\cdots$ Metal began to tear through the thin wings outside the small windows and pinged heavily off the underside of the plane, a quiet arrival in Poland wasn't going to be an option anymore.}\\
        \midrule
        \textsf{\textbf{Persona A:} [Boruc]~A \hlred{skilled} pilot, \hlred{trained} in operating flight controls on the most common planes.}\\
        \textsf{\textbf{Generated Story by \textsc{ConPer} Conditioned on Persona A:} $\cdots$He had been doing this much before, almost a week ago. \hlred{He took a long pull of the airlock, checked his controls, and made a quick mental note of the exact sequence of instructions.} He knew that he couldn't be sure if this would be safe for much longer$\cdots$}\\
        \midrule
        \textsf{\textbf{Persona B:} [Boruc]~An \hlred{unskilled} pilot, and \hlred{never trained} in operating flight controls.}\\
        \textsf{\textbf{Generated Story by \textsc{ConPer} Conditioned on Persona B:} $\cdots$
        He cursed as the plane suffered a complete failure and in a way had caused it to come to a stop, $\cdots$
        \hlred{He'd never flown before, so he didn't know how to pilot in this situation and his experience of the controls had not been good either}$\cdots$}\\
    \bottomrule
\end{tabular}
\caption{An example for controlling the protagonist's persona in story generation. The \textbf{Context} and \textbf{Persona A} are sampled from the \textsc{storium} dataset~\cite{akoury2020storium}. The protagonist's name is shown in the square bracket. And we manually write \textbf{Persona B} based on \textbf{Persona A}. We highlight the sentences which embody the given personas in \hlred{red}.}
\label{tab:example_story}
\end{table}

Stories are important for entertainment. They are made engaging often by portraying animated and believable characters since a story plot unfolds as the characters interact with the object world created in the story~\cite{young2000creating}. Cognitive psychologists determined that the ability of an audience to comprehend a story is strongly correlated with the characters' believability~\cite{graesser1991question}. And the believability mostly depends on whether the characters' reaction to what has happened and their deliberate behavior accord with their personas~(e.g., weakness, abilities, occupations)~\cite{madsen2009exploring,riedl2010narrative}.
Furthermore, previous studies have also stressed the importance of personas in stories to maintain the interest of audience and instigate their sense of empathy and relatedness~\cite{cavazza2009emotional,chandu2019my}. However, despite the broad recognition of its importance, it has not yet been widely explored to endow characters with specified personalities in story generation.

In this paper, we present the first study to impose free-form controllable persona on story generation. Specifically, we require generation models to generate a coherent story, where the protagonist should exhibit the desired personality. We focus on controlling the persona of only the protagonist of a story in this paper and leave the modeling of personas of multiple characters for future work. As exemplified in Table~\ref{tab:example_story}, given a context to present the story settings including characters, location, problems ~(e.g., {``Boruc''} was suffering from a plane crash) and a persona description, the model should generate a coherent story to exhibit the persona~(e.g., what happened when {``Boruc''} was {``skilled''} or {``unskilled''}). In particular, we require the model to embody the personality of the protagonist implicitly through his actions ~(e.g., {``checked his controls''} for the personality {``skilled pilot''}). Therefore, the modeling of relations between persona and events is the \textbf{first challenge} of this problem. Then, we observe that only a small amount of events in a human-written story relate to personas directly and the rest serve for explaining the cause and effect of these events to maintain the coherence of the whole story. Accordingly, the \textbf{second challenge}  is learning to plan a coherent event sequence (e.g., first {``finding the plane shaking''}, then {``checking controls''}, and finally {``landing safely''}) to embody personas naturally.

In this paper, we propose a generation model named {\textsc{ConPer}} to deal with {\textit{Con}}trolling  {\textit{Per}}sona of the protagonist in story generation. Due to the persona-sparsity issue that most events in a story do not embody the persona, directly fine-tuning on real-world stories may mislead the model to focus on persona-unrelated events and regard the persona-related events as noise~\cite{zheng2020pre}. Therefore, before generating the whole story, \textsc{ConPer} first plans persona-related events through predicting one target sentence, which should be motivated by the given personality following the leading context. To this end, we extract persona-related events that have a high semantic similarity with the persona description in the training stage. Then, \textsc{ConPer} plans the  plot as a sequence of keywords to complete the cause and effect of the predicted persona-related events with the guidance of commonsense knowledge. Finally, \textsc{ConPer} generates the whole story conditioned on the planned plot. The stories are shown to have better coherence and persona-consistency than state-of-the-art baselines. 

We summarize our contributions as follows: 

\noindent\textbf{I.} We propose a new task of controlling the personality of the protagonist in story generation.

\noindent\textbf{II.} We propose a generation model named \textsc{ConPer} to impose specified persona into story generation by planning persona-related events and a keyword sequence as intermediate representations.

\noindent\textbf{III.} We empirically show that \textsc{ConPer} can achieve better controllability of persona and generate more coherent stories than strong baselines.


\section{Related Work}
\paragraph{Story Generation} There have been wide explorations for various story generation tasks, such as story ending generation~\cite{guan2019story}, story completion~\cite{DBLP:conf/ijcai/Wang019b} and story generation from short prompts~\cite{fan2018hierarchical}, titles~\cite{yao2019plan} or beginnings~\cite{guan2020knowledge}. To improve the coherence of story generation, prior studies usually first predicted intermediate representations as plans and then generated stories conditioned on the plans. The plans could be a series of keywords~\cite{yao2019plan}, an action sequence~\cite{fan2019strategies, DBLP:conf/emnlp/Goldfarb-Tarrant20} or a keyword distribution~\cite{kang2020plan}. In terms of character modeling in stories, some studies focused on learning characters' persona as latent variables~\cite{bamman2013learning,bamman2014bayesian} or represented characters as learnable embeddings~\cite{ji2017dynamic,clark2018neural,liu2020character}.  
\citet{chandu2019my} proposed five types of specific personas for visual story generation. 
\citet{brahman2021let} formulated two new tasks including character description generation and character identification.
In contrast, we focus on story generation conditioned on personas in a free form of text to describe one's strengths, weaknesses, abilities, occupations and goals.

\paragraph{Controllable Generation}

\begin{figure*}[!ht]
\includegraphics[width=\linewidth]{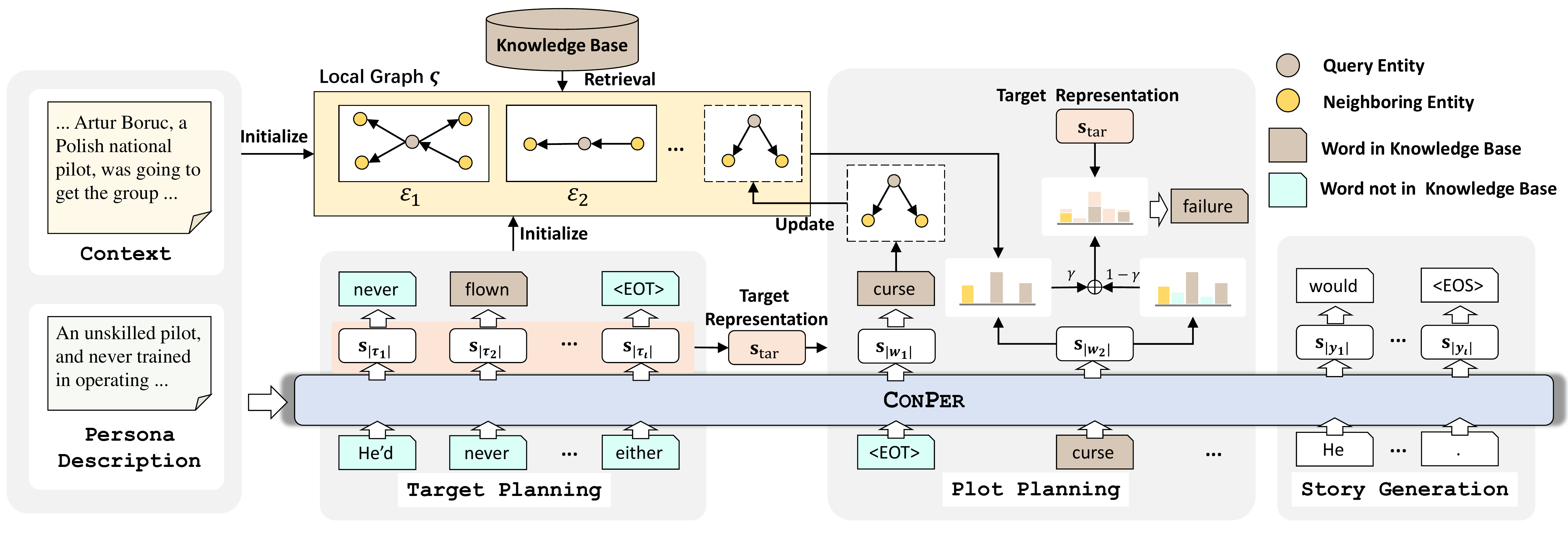}
  \caption{Model overview of \textsc{ConPer}.  The training process is divided into the following three stages: (a) Target Planning: planning persona-related events (called ``target'' for short); (b) Plot Planning: planning a keyword sequence as an intermediate representation of the story with the guidance of the target and a dynamically growing local knowledge graph; And (c) Story Generation: generating the whole story conditioned on the input and plans.}
  \label{tab:model}
\end{figure*}

Controllable text generation aims to generate texts with specified attributes. For example, \citet{DBLP:journals/corr/abs-1909-05858} pretrained a language model conditioned on control codes of different attributes~(e.g., domains, links). \citet{DBLP:conf/iclr/DathathriMLHFMY20} 
proposed to combine a pretrained language model with trainable attribute classifiers to 
increase the likelihood of the target attributes.
Recent studies in dialogue models focused on controlling through sentence functions~\cite{ke2018generating}, politeness~\cite{niu2018polite} and conversation targets~\cite{tang2019target}. For storytelling, \citet{brahman2020cue} incorporated additional phrases to guide the story generation. \citet{brahman2020modeling} proposed to control the emotional trajectory in a story by regularizing the generation process with reinforcement learning. \citet{rashkin2020plotmachines} generated stories from outlines of characters and events by tracking the dynamic plot states with a memory network.

A similar research to ours is \citet{zhang2018personalizing}, which introduced the PersonaChat dataset for endowing the chit-chat dialogue agents with a consistent persona. However, dialogues in PersonaChat tend to exhibit the given personas explicitly~(e.g., the agent says ``I am terriﬁed of dogs'' for the persona ``I am afraid of dogs''). For quantitative analysis, we compute the ROUGE score~\cite{lin-2004-rouge} between the persona description and the dialogue or story. We find that the rouge-2 score is 0.1584 for  {PersonaChat} and 0.018 for our dataset~(i.e., \textsc{storium}). The results indicate that exhibiting personas in stories requires a stronger ability to associate the action of a character and his implicit traits compared with exhibiting personas in dialogues.

\paragraph{Commonsense Knowledge} Recent studies have demonstrated that incorporating external commonsense knowledge significantly improved the coherence and informativeness for dialog generation~\cite{DBLP:conf/ijcai/ZhouYHZXZ18,DBLP:journals/corr/abs-2012-08383}, story ending generation~\cite{guan2019story}, essay generation~\cite{DBLP:conf/acl/YangLLLS19}, story generation~\cite{guan2020knowledge,DBLP:conf/emnlp/XuPSPFAC20,mao2019improving} and story completion~\cite{ammanabrolu2020automated}. These studies usually retrieved a static local knowledge graph which contains entities mentioned in the input, and their related entities. We propose to incorporate the knowledge dynamically during generation to better model the
keyword transition in a long-from story.

\section{Methodology}

We define our task as follows: given a context ${X}=(x_1,x_2,\cdots,x_{|X|})$ with $|X|$ tokens, and a persona description for the protagonist ${P}=(p_1,p_2,\cdots,p_l)$ of length $l$, the model should generate a coherent story ${Y}=(y_1,y_2,\cdots,y_{|Y|})$ of length $|Y|$ to exhibit the persona. To tackle the problem, the popular generation model such as GPT2 commonly employ a left-to-right decoder to minimize the negative log-likelihood $\mathcal{L}_{ST}$ of human-written stories:
\begin{align}
    \mathcal{L}_{ST}&=-\sum_{t=1}^{|Y|}\text{log}P(y_t|y_{<t},S),\\
    P(y_t|y_{<t},S)&=\text{softmax}(\textbf{s}_t\boldsymbol{W}+\boldsymbol{b}),\\
    \textbf{s}_t&=\texttt{Decoder}(y_{<t}, S),
\end{align}
where $S$ is the concatenation of $X$ and $P$, $\textbf{s}_t$ is the decoder's hidden state at the $t$-th position of the story, $\boldsymbol{W}$ and $\boldsymbol{b}$ are trainable parameters. Based on this framework, we divide the training process of \textsc{ConPer} into three stages as shown in Figure~\ref{tab:model}.

\subsection{Target Planning}
We observe that most sentences in a human-written story do not aim to exhibit any personas, but serve to maintain the coherence of the story. Fine-tuning on these stories directly may mislead the model to regard input personas as noise and focus on modeling the persona-unrelated events which are in the majority. Therefore, we propose to first predict persona-related events~(i.e., the target) before generating the whole story.

We use an automatic approach to extract the target from a story since there is no available manual annotation. Specifically, we regard the sentence as the target which has the highest semantic similarity with the persona description. We consider only one sentence as the target in this work due to the persona-sparsity issue, and we also present the result of experimenting with two sentences as the target in the appendix~\ref{appendix_target}. More explorations of using multiple target sentences are left as future work. We adopt NLTK~\cite{bird2009natural} for sentence tokenization. And we measure the similarity between sentences using BERTScore$_{\rm Recall}$~\cite{zhang2019bertscore} with RoBERTa$_{\rm Large}$~\cite{liu2019roberta} as the backbone model. Let $T=(\tau_1,\tau_2,\cdots,\tau_\iota)$ denote the target sentence of length $\iota$, which should be a sub-sequence of $Y$. Formally, the loss function $\mathcal{L}_{TP}$ for this stage can be derived as follows:
\begin{align}
    \mathcal{L}_{TP}&=-\sum_{t=1}^{\iota}\text{log}P(\tau_t|\tau_{<t}, S).
\end{align}
In this way, we exert explicit supervision to encourage the model to condition on the input personas. 

\subsection{Plot Planning}
At this stage, \textsc{ConPer} learns to plan a keyword sequence for subsequent story generation~\cite{yao2019plan}. Plot planning requires a strong ability to model the causal and temporal relationship in the context for expanding a reasonable story plot~(e.g., associating \textit{``unskilled''} with \textit{``failure''} for the example in Table~\ref{tab:example_story}), which is extremely challenging without any external guidance, for instance, commonsense knowledge. In order to plan a coherent event sequence,  we introduce a dynamically growing local knowledge graph, a subset of the external commonsense knowledge base ConceptNet~\cite{speer2017conceptnet}, which is initialized to contain triples related to the keywords mentioned in the input and target. When planning the next keyword, \textsc{ConPer} combines the knowledge information from the local graph and the contextualized features captured by the language model with learnable weights. Then \textsc{ConPer} grows the local graph by adding the knowledge triples neighboring the predicted keyword. Formally, we denote the keyword sequence as ${W}=(w_{1}, w_2,\cdots, w_{k})$ of length $k$ and the local graph as $\mathcal{G}_t$ for predicting the keyword $w_{t}$. The loss function $\mathcal{L}_{KW}$ for generating the keyword sequence is as follows:
\begin{align}
    \mathcal{L}_{KW}&=-\sum_{t=1}^{k}\text{log}P(w_t|w_{<t}, S, T, \mathcal{G}_{t}).
\end{align}

\paragraph{Keyword Extraction}

We extract words that relate to emotions and events from each sentence of a story as keywords for training, since they are important for modeling characters' evolving psychological states and their behavior. We measure the emotional tendency of each word using the sentiment analyzer in NLTK, which predicts a distribution over four basic emotions, i.e., \textit{negative}, \textit{neutral}, \textit{positive}, and \textit{compound}. We regard those words as related to emotions whose scores for \textit{negative} or \textit{positive} are larger than $0.5$. Secondly, we extract and lemmatize the nouns and verbs~(excluding stop-words) from a story as {event-related} keywords with NLTK for POS-tagging and lemmatization. Then we combine the two types of keywords in the original order as the keyword sequence for planning. We limit the number of keywords extracted from each sentence in stories up to 5, and we ensure that there is at least one keyword for a sentence by randomly choosing one word if no keywords are extracted. We don't keep this limitation when extracting keywords from the leading context and the persona description, since these keywords are only used to initialize the local knowledge graph. 

\paragraph{Incorporating Knowledge} We introduce a dynamically growing local knowledge graph for plot planning. For each example, we initialize the graph $\mathcal{G}_1$ as a set of knowledge triples where the keywords in $S$ and $T$ are the head or tail entities, and then update $\mathcal{G}_{t}$ to $\mathcal{G}_{t+1}$ by adding triples related with the generated keyword $w_t$ at $t$-th step. Then, the key problem at this stage is representing and utilizing the local graph for next keyword prediction.

The local graph consists of multiple sub-graphs, each of which contains all the triples related with a keyword denoted as $\varepsilon_i=\{(h^i_n,r^i_n,t^i_n)|h^i_n\in\mathcal{V}, r^i_n\in\mathcal{R}, t^i_n\in\mathcal{V}\}\}|_{n=1}^N$, 
where $\mathcal{R}$ and $\mathcal{V}$ are the relation set and entity set of ConceptNet, respectively. We derive the representation $\textbf{g}_i$ for $\varepsilon_i$ using graph attention~\cite{zhou2018commonsense} as follows:
\begin{align}
\textbf{g}_i&=\sum_{n=1}^{N}\alpha_n[\textbf{h}^i_n;\textbf{t}^i_n]\\
\alpha_n&=\frac{\text{exp}(\beta_n)}{\sum_{j=1}^{N}\text{exp}(\beta_j)},\\
\beta_n&=(\boldsymbol{W}_r\textbf{r}^i_n)^T\text{tanh}(\boldsymbol{W}_h\textbf{h}^i_n+\boldsymbol{W}_t\textbf{t}^i_n),
\end{align}
where $\boldsymbol{W}_h,\boldsymbol{W}_r$ and $\boldsymbol{W}_t $ are trainable parameters, $\textbf{h}^i_n, \textbf{r}^i_n$ and $\textbf{t}^i_n$ are learnable embedding representations for $h^i_n, r^i_n$ and $t^i_n$, respectively. 
We use the same BPE tokenizer~\cite{radford2019language} with the language model to tokenize the head and tail entities, which may lead to multiple sub-words for an entity. 

Therefore, we derive $\textbf{h}^i_n$ and $\textbf{t}^i_n$ by adding the embeddings of all the sub-words. And we initialize the relation embeddings randomly. 

After obtaining the graph representation, we predict the distribution of the next keyword by dynamically deciding whether to select the keyword from the local graph as follows:
\begin{align}
P(w_t|w_{<t}, S, T, \mathcal{G}_{t})&=\gamma_t P^t_{k} + (1-\gamma_t)P^t_{l},
\end{align}
where $\gamma_t\in\{0,1\}$ is a binary learnable weight,  and $P^t_{l}$ is a distribution over the whole vocabulary while $P^t_{k}$ is a distribution over the entities in $\mathcal{G}_t$.
We incorporate the knowledge information implicitly for computing both distributions:
\begin{align}
P^t_{k} &= \text{softmax}(\boldsymbol{W}_k[\textbf{s}_t; \textbf{c}_t]+\boldsymbol{b}_k),\label{ptk}\\
P^t_{l} &= \text{softmax}(\boldsymbol{W}_l[\textbf{s}_t; \textbf{c}_t]+\boldsymbol{b}_l),\label{ptl}
\end{align}
where $\boldsymbol{W}_k,\boldsymbol{b}_k,\boldsymbol{W}_p$ and $\boldsymbol{b}_p$ are trainable parameters, and $\textbf{c}_t$ is a summary vector of the knowledge information by attending on the representations of all the sub-graphs in $\mathcal{G}_t$, formally as follows:
\begin{align}
\textbf{c}_t&=\sum_{n=1}^N\alpha_n\textbf{g}_n,\\
\alpha_n&=\text{softmax}(\textbf{s}_t^T\boldsymbol{W}_g\textbf{g}_n).
\end{align}
where $\boldsymbol{W}_g$ is a trainable parameter. During training process, we set $\gamma_t$ to the ground-truth label $\hat{\gamma}_t$. During generation process, we decide $\gamma_t$ by deriving the probability $p_t$ of selecting an entity from the local graph as the next keyword. And we set $\gamma_t$ to $1$ if $p_t<0.5$ otherwise 0. We compute $p_t$  as follows:
\begin{align}
p_t &= \text{sigmoid}(\boldsymbol{W}_p[\textbf{s}_t; \textbf{c}_t]+\boldsymbol{b}_p),
\end{align}
where $\boldsymbol{W}_p$ and $\boldsymbol{b}_p$ are trainable parameters. We train the classifier with the standard cross entropy loss $\mathcal{L}_{C}$ derived as follows:
\begin{align}
    \mathcal{L}_{C}=-\big(\hat{\gamma}_t\text{log}p_t+(1-\hat{\gamma}_t)\text{log}(1-{p}_t)\big),
\end{align}
where $\hat{\gamma}_t$ is the ground-truth label. In summary, the overall loss function $\mathcal{L}_{PP}$ for the plot planning stage is computed as follows:
\begin{align}
    \mathcal{L}_{PP}=\mathcal{L}_{KW}+\mathcal{L}_{C}.
\end{align} 
By incorporating commonsense knowledge for planning, and dynamically updating the local graph, \textsc{ConPer} can better model the causal and temporal relationship between events in the context.

\paragraph{Target Guidance} In order to further improve the coherence and the persona-consistency, we propose to exert explicit guidance of the predicted target on plot planning. Specifically, we expect \textsc{ConPer} to predict keywords close to the target in semantics. Therefore, we add a bias term $\textbf{d}^t_k$ and $\textbf{d}^t_l$ into Equation~\ref{ptk} and \ref{ptl}, respectively, formally as follows:
\begin{align}
    P^t_{k}&=\text{softmax}(\boldsymbol{W}_{k}[\textbf{s}_t;\textbf{c}_t]+\boldsymbol{b}_{k}+\textbf{d}^t_k),\\
    \textbf{d}^t_k&=[\textbf{s}_\text{tar};\textbf{c}_t]^T\boldsymbol{W}_d\textbf{E}_k+\boldsymbol{b}_d,\label{bias_term}\\
    \textbf{s}_\text{tar}&=\frac1\iota\sum_{t=1}^\iota\textbf{s}_{|\tau_t|},
\end{align}
where $\boldsymbol{W}_d$ and $\boldsymbol{b}_d$ are trainable parameters, $\textbf{s}_\text{tar}$ is the target representation computed by averaging the hidden states at each position of the predicted target, and $\textbf{E}_k$ is an embedding matrix, each row of which is the embedding for an entity in $\mathcal{G}_t$. The modification for Equation~\ref{ptl} is similar except that we compute the bias term $\textbf{d}_l^t$ with an embedding matrix $\textbf{E}_l$ for the whole vocabulary.

\subsection{Story Generation} 
After planning the target $T$ and the keyword sequence $W$, we train \textsc{ConPer} to generate the whole story conditioned on the input and plans with the standard language model loss $\mathcal{L}_{ST}$. Since we extract one sentence from a story as the target, we do not train \textsc{ConPer} to regenerate the sentence in the story generation stage. And we insert a special token \texttt{Target} in the story to specify the position of the target during training. In the inference time, \textsc{ConPer} first plans the target and plot, then generates the whole story, and finally places the target into the position of \texttt{Target}.

\section{Experiments}

\subsection{Dataset}
We conduct the experiments on the \textsc{storium} dataset~\cite{akoury2020storium}. \textsc{storium} contains nearly 6k long-form stories and each story unfolds through a series of scenes with several shared characters. A scene consists of multiple short scene entries, each of which is written to either portray one character with annotation for his personality~(i.e., the \textit{``card''} in \textsc{storium}), or introduce new story settings~(e.g., problems, locations) from the perspective of the narrator. 
In this paper, we concatenate all entries from the same scene since a scene can be seen as an independent story. And we regard a scene entry written for a certain character as the target output, the personality of the character as the persona description, and the previous entries written for this character or from the perspective of the narrator in the same scene as the leading context. We split the processed examples for training, validation and testing based on the official split of \textsc{storium}. We retain about 1,000 words (with the correct sentence boundary) for each example due to the length limit of the pretrained language model.

At the plot planning stage, we retrieve a set of triples from ConceptNet~\cite{speer2017conceptnet} for each keyword extracted from the input or generated by the model. We only retain those triples of which both the head and tail entity contain one word and occur in our dataset, and the confidence score of the relation~(annotated by ConceptNet) is more than 1.0. The average number of triples for each keyword is 33. We show more statistics in Table~\ref{tab:stat}.

\begin{table}[h]
\small
    \centering
    \begin{tabular}{lccc}
    \toprule
    & \textbf{Train} & \textbf{Valid}  & \textbf{Test} \\
    \midrule
  \textbf{\# Examples}  & 47,910 & 6,477 & 6,063 \\
  \textbf{Avg. Context Length} & 332.7 & 324.8  & 325.7 \\
  \textbf{Avg. Description Length} & 23.8 & 22.7  & 24.6 \\
  \textbf{Avg. Story Length} & 230.5 & 225.7  & 234.3 \\
  \midrule
  \textbf{Avg. Target Length} & 21.8 & 21.7 & 22.2\\
  \textbf{Avg. \# Keywords~(Input)}&101.1&99.2&99.6 \\
    \textbf{Avg. \# Keywords~(Story)}&31.2&30.5&31.5 \\
   \bottomrule
    \end{tabular}
    \caption{Dataset statistics. We compute the length by counting tokens using the BPE tokenizer of GPT2. Keywords are extracted either from the input to initialize the local graph, or from the story to train the model for plot planning.}
    \label{tab:stat}
\end{table}

\subsection{Baselines}
We compare \textsc{ConPer} with following baselines.
\textbf{(1) {ConvS2S:}} It directly uses a convolutional seq2seq model to generate a story conditioned on the input~\cite{gehring2017convolutional}.
\textbf{(2) {Fusion:}} It generates a story by first training a convolutional seq2seq model, and then fixing the model and initializing another trainable convolutional seq2seq model with its parameters. Then the two models are trained together by a fusion mechanism.~\cite{fan2018hierarchical}.
\textbf{(3) {Plan\&Write:} }It first plans a keyword sequence conditioned on the input, and then generates a story based on the keywords~\cite{yao2019plan}. \textbf{(4) {GPT2$_{\rm Scr}$:}} It has the same network architecture with GPT2 but is trained on our dataset from scratch without any pretrained parameters.
\textbf{(5) {GPT2$_{\rm Ft}$:}} It is initialized using pretrained parameters, and then fine-tuned on our dataset with the standard language modeling objective. \textbf{(6) {PlanAhead:}} It first predicts a keyword distribution conditioned upon the input, and then generates a story by combining the language model prediction and the keyword distribution with a gate mechanism~\cite{kang2020plan}. We remove the sentence position embedding and the auxiliary training objective~(next sentence prediction) used in the original paper for fair comparison.

Furthermore, we evaluate the following ablated models to investigate the influence of each component: \textbf{\textsc{(1) ConPer} w/o KG:} removing the guidance of the commonsense knowledge in the plot planning stage. \textbf{\textsc{(2) ConPer} w/o TG:} removing target guidance in the plot planning stage. \textbf{\textsc{(3) ConPer} w/o PP:} removing the plot planning stage, which means the model first plans a target sentence and then directly generates the whole story. \textbf{\textsc{(4) ConPer} w/o TP:} removing the target planning stage, which also leads to the removal of target guidance in the plot planning stage.

\subsection{Experiment Settings}
We build \textsc{ConPer} based on  GPT2~\cite{radford2019language}, which is widely used for story generation~\cite{guan2020knowledge}. We concatenate the context and the persona description with a special token as input for each example. For fair comparison, we also add special tokens at both ends of the target sentence in a training example for all baselines. We implement the non-pretrained models based on the scripts provided by the original papers, and the pretrained models based on the public checkpoints and codes of HuggingFace's Transformers\footnote{\url{https://github.com/huggingface/ transformers}}. And we set all the pretrained models to the base version due to limited computational resources. We set the batch size to 8, the initial learning rate of the AdamW optimizer to 5e-5, and the maximum training epoch to 5 with an early stopping mechanism. And we generate stories using top-$p$ sampling with $p=0.9$~\cite{holtzman2019curious}. We apply these settings to all the GPT-based models, including GPT$_{\rm Scr}$, GPT$_{\rm Ft}$, PlanAhead, \textsc{ConPer} and its ablated models. As for ConvS2S, Fusion and Plan\&Write, we used the settings from their respective papers and codebases.

\begin{table*}[h]
\small
\centering
\begin{tabular}{l|cccc|cccc}
\toprule
\multicolumn{1}{c|}{\multirow{2}{*}{\textbf{Models}}} & \multicolumn{4}{c|}{\textbf{Coherence}}& \multicolumn{4}{c}{\textbf{Persona consistency}}\\
\multicolumn{1}{c|}{}& \textbf{Win(\%)} & \textbf{Lose(\%)} & \textbf{Tie(\%)} & \textbf{$\kappa$} & \textbf{Win(\%)} & \textbf{Lose(\%)} & \textbf{Tie(\%)} & \textbf{$\kappa$} \\ \midrule
\textbf{\textsc{ConPer} vs. ConvS2S}& 89.0* & 5.0 & 6.0          & 0.625 & 82.0* & 8.0 & 10.0 & 0.564\\ 
\textbf{\textsc{ConPer} vs. Fusion}&71.0*&23.0&6.0&0.213&61.0*&22.0&17.0&0.279\\ 
\textbf{\textsc{ConPer} vs. GPT2$_{\rm Ft}$}& 54.0* & 18.0 & 28.0 &     0.275 & 53.0* & 11.0 & 36.0 & 0.215 \\
\textbf{\textsc{ConPer} vs. PlanAhead}& 53.0* & 25.0 & 22.0 &    0.311 & 59.0* & 28.0 & 13.0 & 0.280\\
\bottomrule
\end{tabular}
\caption{Manual evaluation results. The scores indicate the percentage of \textit{win}, \textit{lose} or \textit{tie} when comparing our model with a baseline. $\kappa$ denotes Randolph’s kappa to measure the inter-annotator agreement. * means \textsc{ConPer} outperforms the baseline model significantly with p-value$<0.01$ (Wilcoxon signed-rank test).}
\label{tab:human_eva}
\end{table*}

\begin{table}[!t]
\small
    \centering
    \begin{tabular}{lccccc}
    \toprule
    \textbf{Models}&\textbf{B-1}&\textbf{B-2}&\textbf{BS-t}&\textbf{BS-m}&\textbf{PC}\\
    \midrule

    \textbf{ConvS2S}&12.5&4.7&22.2&32.8&17.1\\
    \textbf{Fusion}&13.3&5.0&22.7&33.3& 30.8\\
    \textbf{Plan\&Write}&7.2&2.8&6.2&29.7&23.6\\
    \midrule
    \textbf{GPT2$_{\rm Scr}$}&13.3&4.8&24.7&38.0&26.6\\
    \textbf{GPT2$_{\rm Ft}$}&13.5&4.7&26.7&37.8&39.5\\
    \textbf{PlanAhead}&15.4&5.3&26.1&37.8&50.2\\
    \midrule
    \textbf{\textsc{ConPer}}&\textbf{19.1}&\textbf{6.9}&\textbf{32.1}&\textbf{41.4}&\textbf{59.7}\\
    \textbf{~~w/o KG}&17.4&\underline{6.3}&31.6&39.7&53.4\\ \textbf{~~w/o TG}&\underline{17.7}&\underline{6.3}&31.9&\underline{40.2}&\underline{56.3}\\
    \textbf{~~w/o PP}&14.9&5.3&\underline{32.0}&40.0&46.9\\
    \textbf{~~w/o TP}&16.4&5.8&27.8&37.7&44.9\\
    \midrule
    \textbf{\textit{Grouth Truth}}&\textit{N/A}&\textit{N/A}&\textit{42.6}&\textit{42.6}&\textit{75.2}\\
   \bottomrule
    \end{tabular}
    \caption{Automatic evaluation results. The best performance is highlighted in \textbf{bold}, and the second best is \underline{underlined}. All results are multiplied by 100.}
    \label{tab:auto_eval}
\end{table}

\subsection{Automatic Evaluation}
\paragraph{Metrics} We adopt the following automatic metrics for evaluation on the test set. \textbf{(1) BLEU~(\textbf{B-n}):} We use $n=1,2$ to evaluate $n$-gram overlap between generated and ground-truth stories~\cite{papineni2002bleu}. \textbf{(2) BERTScore-target~(BS-t):} We use BERTScore$_{\rm Recall}$~\cite{zhang2019bertscore} to measure the semantic similarity between the generated target sentence and the persona description. A higher result indicates the target embodies the persona better. \textbf{(3) BERTScore-max~(BS-m)}: It computes the maximum value of BERTScore between each sentence in the generated story and the persona description. \textbf{(4) Persona-Consistency~(PC):} It is a learnable automatic metric~\cite{guan2020union}. We fine-tune RoBERTa$_{\textsc{base}}$ on the training set as a classifier to distinguish whether a story exhibits a consistent persona with a persona description. We regard the ground-truth stories as positive examples where the stories and the descriptions are consistent, and construct negative examples by replacing the story with a randomly sampled one. After fine-tuning, the classifier achieves an 83.63\% accuracy on the auto-constructed test set. Then we calculate the consistency score as the average classifier score of all the generated texts regarding the corresponding input. 

\paragraph{Result}
Table \ref{tab:auto_eval} shows the automatic evaluation results. \textsc{ConPer} can generate more word overlaps with ground-truth stories as shown by higher BLEU scores. And \textsc{ConPer} can better embody the specified persona in the target sentence and the whole story as shown by the higher BS-t and BS-m score. The higher PC score of \textsc{ConPer} also further demonstrate the better exhibition of given personas in the generated stories. As for ablation tests, all the ablated models have lower scores in terms of all metrics than \textsc{ConPer}, indicating the effectiveness of each component. Both \textsc{ConPer} w/o PP and \textsc{ConPer} w/o TP drop significantly in BLEU scores, suggesting that planning is important for generating long-form stories.
\textsc{ConPer} w/o TP also performs substantially worse in all metrics than \textsc{ConPer} w/o TG, indicating the necessity of explicitly modeling the relations between persona descriptions and story plots. We also show analysis of target guidance in Appendix \ref{appendix:target_guide}.

\subsection{Manual Evaluation}
We conduct a pairwise comparison between our model and four strong baselines including PlanAhead, GPT2$_{{\rm Ft}}$, Fusion and ConvS2S. We randomly sample 100 stories from the test set, and obtain 500 stories generated by \textsc{ConPer} and four baseline models. For each pair of stories (one by \textsc{ConPer}, and the other by a baseline, along with the input), we hire three annotators to give a preference (\textit{win}, \textit{lose} or \textit{tie}) in terms of \textit{coherence}~(inter-sentence relatedness, causal and temporal dependencies) and \textit{persona-consistency} with the input (exhibiting consistent personas). We adopt majority voting to make the final decisions among three annotators. Note that the two aspects are independently evaluated. We resort to Amazon Mechanical Turk~(AMT) for the annotation.  As shown in Table \ref{tab:human_eva}, \textsc{ConPer} outperforms baselines significantly in coherence and persona consistency. 

\begin{table}[!h]
\small
\centering
\begin{tabular}{l|ccc}
\toprule
\textbf{{Policies}}& \textbf{Yes~(\%)} & \textbf{No~(\%)} &  \textbf{$\kappa$} \\ 
\midrule
\textbf{Random}& 22.0 & 78.0 & 0.25\\ 
\textbf{Ours}& 75.0 & 25.0 & 0.36\\ 
\bottomrule
\end{tabular}
\caption{Percentages of examples labeled with ``Yes'' or ``No'' for whether the identified sentence reflects the given persona. $\kappa$ denotes Randolph’s kappa to measure the inter-annotator agreement.}
\label{tab:human_eva_target}
\end{table}

Furthermore, we used human annotation to evaluate whether the identified target sentence embodies the given persona. We randomly sampled 100 examples from the test set, and identified the target for each example as the sentence with the maximum BERTScore with the persona description. And we used a random policy as a baseline which randomly samples a sentence from the original story as the target. We hired three annotators on AMT to annotate each example~(``Yes'' if the sentence embodies the given persona, and ``No'' otherwise). We adopted majority voting to make the final decision among three annotators. Table \ref{tab:human_eva_target} shows our method significantly outperforms the random policy in identifying the persona-related sentences.

\subsection{Controllability Analysis}
To further investigate whether the models can be generalized to generate specific stories to exhibit different personas conditioned on the same context, we perform a quantitative study to observe how many generated stories are successfully controlled as the input persona descriptions change. 
\paragraph{Automatic Evaluation} For each example in the test set, we use a model to generate ten stories conditioned on the context of this example and ten persona descriptions randomly sampled from other examples, respectively. 
We regard a generated story as successfully controlled if the pair of the story and its corresponding persona description~(along with the context) has the maximum persona-consistency score among all the ten descriptions. 
We regard the average percentages of the stories which are successfully controlled in all the ten generated stories for each example in the whole test set as the controllability score of the model. We show the results for \textsc{ConPer} and strong baselines in Table~\ref{tab:auto_controbility}. Furthermore, we also compute the superiority~(denoted as $\Delta$) of the persona-consistency score computed between a generated story and its corresponding description compared to that computed between the story and one of the other nine descriptions~ \cite{sinha-etal-2020-learning}. A larger $\Delta$ means the model can generate more specific stories adhering to the personas. 

\begin{table}[!t]
\small
    \centering
    \begin{tabular}{lcc}
    \toprule
    \textbf{Models}&\textbf{Controllability Score}&\textbf{$\Delta$}\\
    \midrule
    \textbf{Plan\&Write}&10.6&0.01\\
    \textbf{GPT2$_{\rm Ft}$}&24.2&11.2\\
    \textbf{PlanAhead}&23.1&11.2\\
    \midrule
    \textbf{\textsc{ConPer}}&\textbf{29.5}&\textbf{15.1}\\
   \bottomrule
    \end{tabular}
    \caption{Automatic evaluation results for the controllability. All results are mulitplied by 100.}
    \label{tab:auto_controbility}
\end{table}

\begin{table}[!t]
\small
    \centering
    \begin{tabular}{lccc}
    \toprule
    \textbf{Models}&\textbf{Acco (\%)}&\textbf{Oppo (\%)}&\textbf{Irre (\%)}\\
    \midrule
    \textbf{GPT2$_{\rm Ft}$}&21&10&69\\
    \textbf{PlanAhead}&44&12&44\\
    \midrule
    \textbf{\textsc{ConPer}}&\textbf{66}&9&25\\
   \bottomrule
    \end{tabular}
    \caption{Manual evaluation results for the controllability. Acco/Oppo/Irre means the example exhibits an accordant/opposite/irrelevant persona with the input.}
    \label{tab:manual_controbility}
\end{table}

As shown in Table \ref{tab:auto_controbility}, there are more stories successfully controlled for \textsc{ConPer} than baselines. And the larger $\Delta$ of \textsc{ConPer} suggests that it can generate more specific stories to the input personas. The results show the better generalization ability of \textsc{ConPer} to generate persona-controllable stories.

\paragraph{Manual Evaluation} 
For manual evaluation, we randomly sampled 50 examples from the test set, and manually revised the persona descriptions to exhibit an opposite persona~(e.g., from ``skilled pilot'' to ``unskilled pilot''). We required a model to generate two stories conditioned on the original and its opposite persona description, respectively. Finally we obtained 300 stories from three models including GPT2$_{\rm Ft}$, PlanAhead and \textsc{ConPer}. Then, we hired three graduates to judge whether each story accords with the input persona. All annotators have good English language proficiency and are well trained for this evaluation task. 
Table~\ref{tab:manual_controbility} shows the evaluation results. We can see that 66\% of the stories generated by \textsc{ConPer} are accordant with the input persona, suggesting the better controllability of \textsc{ConPer}.

\begin{table*}[h]
\small
\begin{tabular}{p{443pt}}
    \toprule
        \textbf{Context:} 
        $\cdots$ the \underline{group} has \underline{gathered} on the rooftop \underline{garden} of Miyamoto Mansion $\cdots$ the TV \underline{set} out \underline{near} the long \underline{table} on the \underline{patio} is \underline{talking} about some \underline{spree} of \underline{thefts} at \underline{low volume} $\cdots$ the \underline{issue} of Chloe's \underline{disappearance} and the \underline{missing statue} still \underline{hang} over their heads. \\
        \\
        \textbf{Persona Description:} [Aito]~You are above \underline{average} in your \underline{computer skills}. 
        If \underline{information} is \underline{power}, then your \underline{ability} to \underline{use} the internet
        \underline{makes} you one of the most \underline{powerful people} on the \underline{planet}.\\
        \midrule
        \textbf{GPT2$_{{\rm Ft}}$:}
        {\textit{Aito looked at the others, still trying to help find a way out of the hotel.  He wasn't sure what the rest of the group wanted to see if they were going to survive and all knew if he needed to be needed $\cdots$}}\\
        \midrule
        \textbf{PlanAhead:}
        {Miyamoto Mansion $\cdots$ perhaps it's just a bit farther away. The music sounds bright enough but the line of visitors does not. \textit{Aito was once a pretty girl, he had always been quite witty when talking to people but she always found it annoying that a group of tourists looked like trash just to her} $\cdots$ }\\
    \midrule
    \textbf{\textsc{ConPer}:}
     {$\cdots$ ``Oh, wait $\cdots$ wait $\cdots$ people are talking about Chloe?'' $\cdots$  ``\textbf{\hlred{I have a feeling the internet is probably our best chance to get through this}}'' $\cdots$ Aito looked around the table a moment before \hlred{pulling out her tablet and starting typing furiously into her computer}. She looked up at the tablet that had appeared, and she could see that it was working on a number of things$\cdots$}\\\\
    \textbf{{Planned keywords}:} $\cdots$ people $\rightarrow$ look $\rightarrow$ around $\rightarrow$ tablet $\rightarrow$ see $\cdots$\\
    \bottomrule
\end{tabular}
\caption{Generated stories by different models. \textit{Italic} words indicate the improper entities or events in terms of the consistency with the input. The \textbf{bold} sentence indicate the generated target by \textsc{ConPer}. \hlred{Red} words denote the consistent events adhering to the input. And the extracted keywords are \underline{underlined}.}
\label{tab:case_study}
\end{table*}

\subsection{Case Study}
We present some cases in Table \ref{tab:case_study}. We can see that the story generated by \textsc{ConPer} exhibits the specified persona with a coherent event sequence. The planned keywords by \textsc{ConPer} provide an effective discourse-level guidance for the subsequent story generation, such as \texttt{tablet}, which has a commonsense connection with \texttt{computer skills} and \texttt{Internet} in the input. In contrast, the baselines tend to not generate any persona-related events. For example, the given persona description emphasizes the strong computer skills of the protagonist while the stories generated by PlanAhead and GPT2 have nothing to do with the computer skills. We further analyze some error cases generated by our model in Appendix \ref{appendix:error}.

\section{Conclusion}
We present \textsc{ConPer}, a planning-based model for a new task aiming at controlling the protagonist's persona in story generation. We propose target planning to explicitly model the relations between persona-related events and input personas, and plot planning to learn the keyword transition in a story with the guidance of predicted persona-related events and external commonsense knowledge. Extensive experiments show that \textsc{ConPer} can generate more coherent stories with better consistency with the input personas than strong baselines. Further analysis also indicates the better persona-controllability of \textsc{ConPer}.

\section*{Acknowledgement}
This work was supported by the National Science Foundation for Distinguished Young Scholars (with No. 62125604) and the NSFC projects (Key project with No. 61936010 and regular project with No. 61876096). This work was also supported by the
Guoqiang Institute of Tsinghua University, with Grant No. 2019GQG1 and 2020GQG0005. This work was also sponsored by Tsinghua-Toyota Joint Research Fund. We would also like to thank the anonymous reviewers for their invaluable suggestions and feedback.

\section*{Ethics Statements}
We conduct the experiments by adapting a public story generation dataset \textsc{storium} to our task. Automatic and manual evaluation results show that our model \textsc{ConPer} outperforms existing state-of-the-art models in terms of coherence, consistency and controllability, suggesting the generalization ability of \textsc{ConPer} to different input personas. 
And our approach can be easily extended to different syntactic levels~(e.g., phrase-level and paragraph-level events), different model architectures~(e.g., BART~\cite{bart}) and different generation tasks~(e.g., stylized long text generation).

In both \textsc{storium} and ConceptNet, we find some potentially offensive words. Therefore, our model may suffer from risks of generating offensive content, although we have not observed such content in the generated results. Furthermore, ConceptNet consists of commonsense triples of concepts, which may not be enough for modeling inter-event relations in long-form stories. 
We resort to Amazon Mechanical Turk (AMT) for manual evaluation. We do not ask about personal privacy or collect personal information of annotators in the annotation process. We hire three annotators and pay each annotator \$0.1 for comparing each pair of stories. The payment is reasonable considering that it would cost average one minute for an annotator to finish a comparison. 
\normalem
\bibliographystyle{acl_natbib}
\bibliography{anthology,acl2022}

\appendix

\section{Implementation Details}
We train our model on one Quadro RTX 6000 GPU. It costs about 25 hours to train our model, and 4 hours to generate stories using our model.

\section{Analysis of Extraction Strategy}

\subsection{Target Extraction}~\label{appendix_target}

We regard one sentence which has the maximum BERTScore with the persona description as the target in our model. We conducted two experiments to further investigate the influence of target extraction strategy: \textbf{\textsc{(1) ConPer}~(Rand):} It regards a sentence randomly sampled from the story as the target for training in the target planning stage. \textbf{\textsc{(2) ConPer}~(Multi)}: It regards two sentences which have the maximum BERTScore with the persona description as the target.

As shown in Table \ref{tab:ablated_auto_eval}, when using a random sentence as the target, all the metrics drop significantly. And Table~\ref{tab:human_eva_target} in the main paper shows that it is hard for the random policy to select persona-related sentences. The results indicate the benefit of our methods for modeling relations between personas and events. Moreover, using multiple sentences as the target is inferior to using only one in terms of most metrics. It is possibly because stories in \textsc{storium} tend to embody personas sparsely, and modeling the relations between personas and multiple persona-unrelated events directly may hurt the performance. The BS-t score is higher when using multiple sentences because more words can easily lead to a higher recall score.

\subsection{Keyword Extraction}

We extracted at most 5 keywords from each sentence for the plot planning stage. We also experimented with a more sparse plan by  extracting only one keyword from each sentence~(called \textbf{\textsc{ConPer}~(Sparse)}). Table \ref{tab:ablated_auto_eval} shows that using a more sparse plan performs worse in all metrics. It is possibly because the limited planning keywords could not make the best of the external knowledge to form coherent and persona-related plots.

\begin{table}[!t]
\small
    \centering
    \begin{tabular}{lccccc}
    \toprule
    \textbf{Models}&\textbf{B-1}&\textbf{B-2}&\textbf{BS-t}&\textbf{BS-m}&\textbf{PC}\\
    \midrule
    \textbf{\textsc{ConPer}}&\textbf{19.1}&\textbf{6.9}&32.1&\textbf{41.4}&\textbf{59.7}\\
    \midrule
    \textbf{\textsc{ConPer}~\scriptsize{(Rand)}}&17.4&6.2&26.0&38.9&52.1\\
    \textbf{\textsc{ConPer}~\scriptsize{(Multi)}}&17.9&6.6&\textbf{32.6}&40.0&55.1\\
    \textbf{\textsc{ConPer}~\scriptsize{(Sparse)}}&18.0&6.6&31.6&40.2&57.0\\
    \midrule
    \textbf{\textit{Grouth Truth}}&\textit{N/A}&\textit{N/A}&\textit{42.6}&\textit{42.6}&\textit{75.2}\\
   \bottomrule
    \end{tabular}
    \caption{Automatic evaluation results for several variants of \textsc{ConPer}. The best performance is highlighted in \textbf{bold}. All results are multiplied by 100.}
    \label{tab:ablated_auto_eval}
\end{table}

\section{Analysis of Target Guidance}\label{appendix:target_guide}

\begin{figure}[!h]
\includegraphics[width=\linewidth]{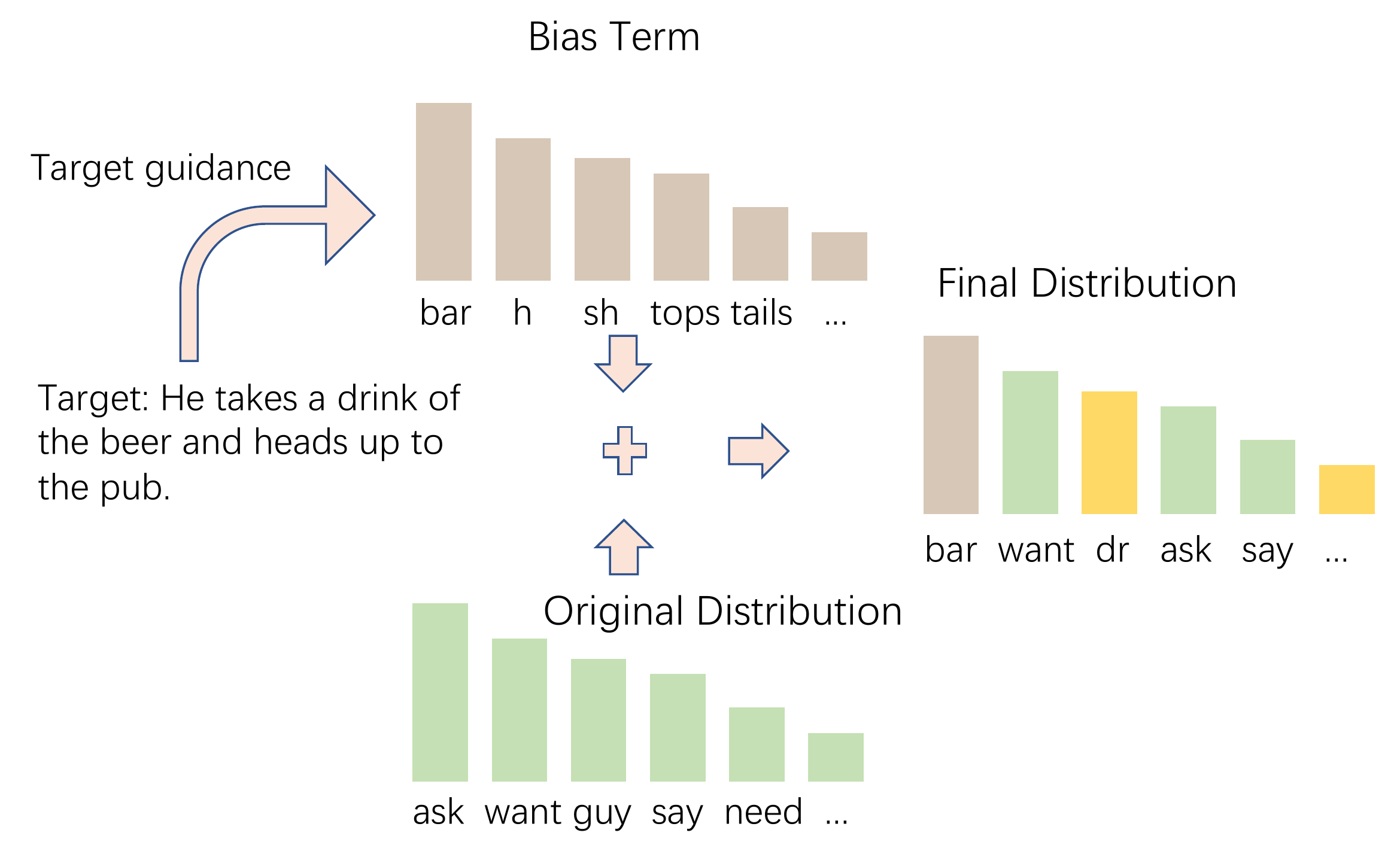}
  \caption{A case showing the effect of target guidance. The planning keywords are brought closer to the target in semantics under the target guidance.}
  \label{tab:target_guidance}
\end{figure}

We visualize how target guidance affects word prediction in the plot planning stage in Figure \ref{tab:target_guidance}. The original word distribution is weighted to those words irrelevant to the target sentence, while the bias term~(Equation \ref{bias_term}) is weighted to those words related to the target sentence in semantics such as \textit{bar}. After combining the original word distribution with the bias term, the final distribution can balance the trade-off between target guidance and language model prediction. This validates our hypothesis that target guidance can draw the planned plots closer to the target, which helps improve the story coherence and persona-consistency.

\section{Diversity}
 We compare the diversity of \textsc{ConPer} with baselines using distinct-$n$~(D-$n$)~\cite{li2016diversity}, the ratio of distinct $n$-grams to all $n$-grams in generated stories. The results in Table~\ref{tab:auto_diversity_eval} show that \textsc{ConPer} has better coherence and persona consistency without sacrificing the diversity.

\begin{table}[h]
\small
    \centering
    \begin{tabular}{lcccc}
    \toprule
    \textbf{Models}&\textbf{D-1}&\textbf{D-2}&\textbf{D-3}&\textbf{D-4}\\
    \midrule
    \textbf{GPT2$_{\rm Scr}$}&0.021&0.134&0.381&0.653\\
    \textbf{GPT2$_{\rm Ft}$}&0.022&0.184&0.501&0.777\\
    \textbf{PlanAhead}&0.032&0.256&0.618&0.863\\
    \textbf{\textsc{ConPer}}&0.016&0.148&0.439&0.730\\
    \midrule
    \textbf{\textit{Grouth Truth}}&\textit{0.062}&\textit{0.368}&\textit{0.739}&\textit{0.927}\\
   \bottomrule
    \end{tabular}
    \caption{Automatic evaluation results. \textsc{ConPer} is comparable with fine-tuned GPT2 in diversity performance.}
    \label{tab:auto_diversity_eval}
\end{table}

\section{Manual Evaluation}
\begin{figure*}[t]
  \centering
  \includegraphics[width=\linewidth]{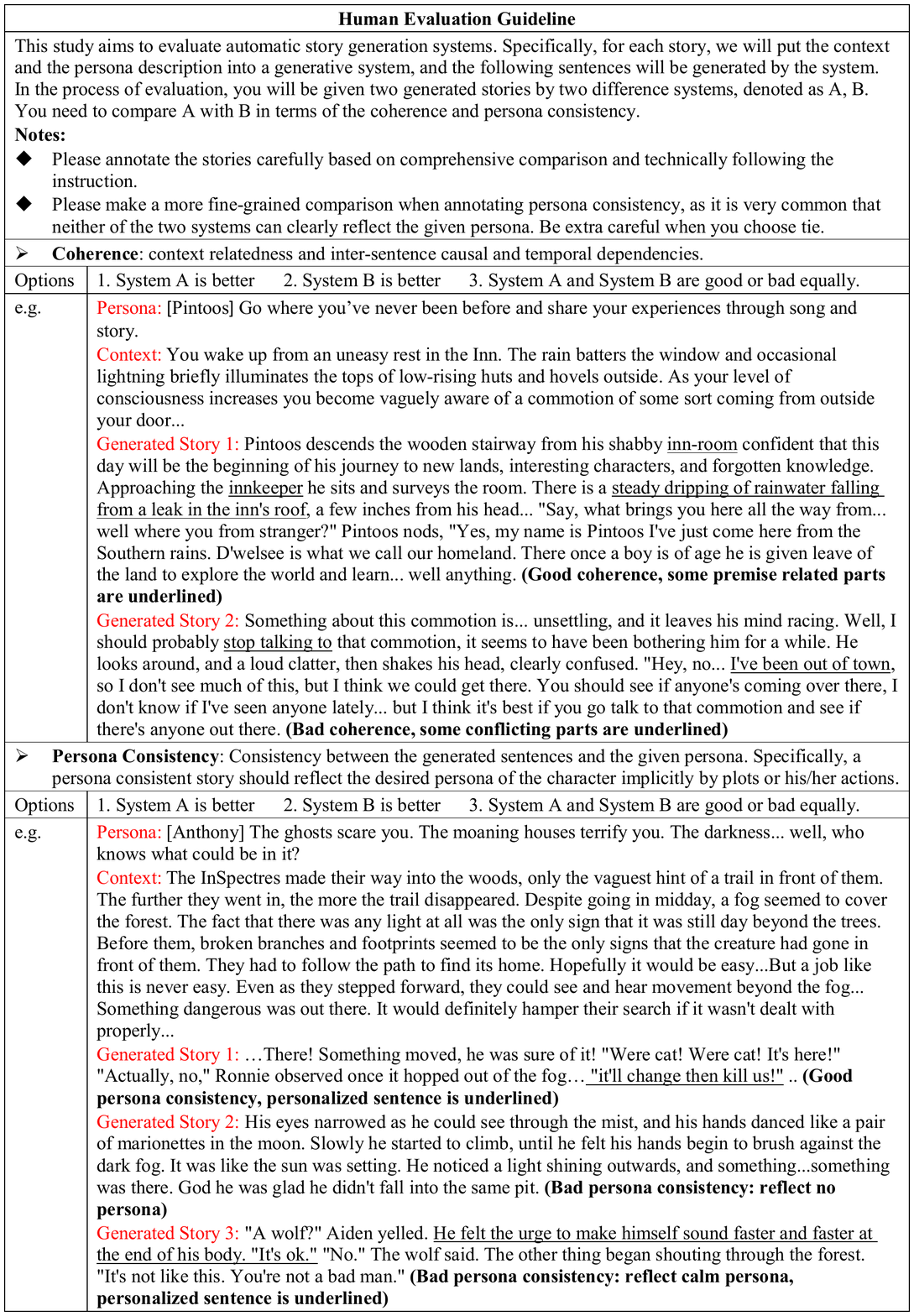}
  \caption{
    The guideline of story quality evaluation.
  }
  \label{fig:guideline}
\end{figure*}

We conduct manual evaluation on Amazon Mechanical Turk. To improve the annotation quality, we provide a detailed instruction for annotators, which contains: (1) a summary of our task; (2) a formal definition for coherence and persona consistency; and (3) good and bad examples for coherence and persona consistency. The detailed evaluation guideline is shown in Figure \ref{fig:guideline}.

\section{Model Parameters}
We compute the number of parameters for some models used in our experiments. The result is shown in Table \ref{tab:parameter}.

\begin{table}[!t]
    \centering
    \begin{tabular}{cc}
    \toprule
      \textbf{Models}   & \textbf{Number of Parameters} \\
      \midrule
      \textbf{ConvS2S}  & 135M\\
      \textbf{Fusion} & 255M\\
      \midrule
      \textbf{GPT2} & 124M\\
      \textbf{PlanAhead}&201M\\
      \midrule
      \textbf{\textsc{ConPer}}&247M\\
      \bottomrule
    \end{tabular}
    \caption{Number of Parameters of different models.}
    \label{tab:parameter}
\end{table}


\section{Error Analysis}\label{appendix:error}

\begin{figure*}[!ht]
  \centering
  \includegraphics[width=0.7\linewidth]{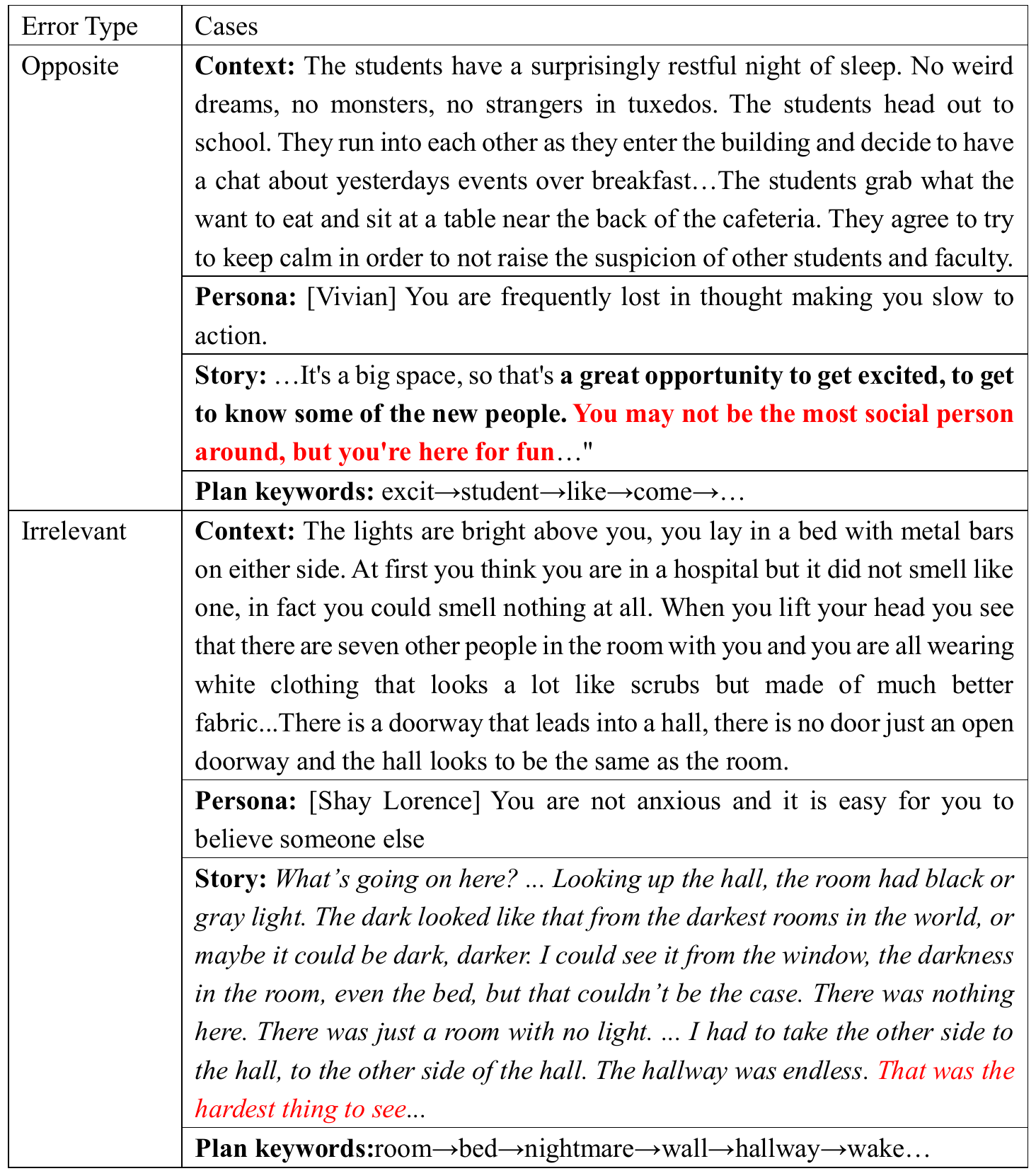}
  \caption{
    Typical errors by our model. \textbf{Bold} words indicate the events exhibiting the opposite persona. \textit{Italic} words indicate the events that are irrelevant with the given persona. And the \textcolor{red}{red} sentence indicate the generated target by \textsc{ConPer}.}
  \label{fig:error}
\end{figure*}

Although the proposed model outperforms the strong baselines, Table \ref{tab:manual_controbility} in the main paper shows that there are still many generated stories that exhibit opposite or irrelevant with the given persona. Therefore, we presented some typical error cases generated by our model for each error type in Figure \ref{fig:error}. These cases show our model still does not completely control personas in story generation. When there is a slight conflict between the generated target sentence and the given persona~(e.g., \texttt{you're here for fun} is slightly conflict with \texttt{slow to action}), the generated plan would further deviate from the input under the guidance of the target sentence~(e.g., \texttt{excit}, \texttt{like}), and finally the generated story exhibits an opposite persona. Similarly, when the generated target sentence is irrelevant with the given persona~(e.g., \texttt{That was the hardest thing too see}), the final generated story doesn't have any persona-related event. These errors also indicate the target sentence plays an important role in controlling the protagonist's persona in story generation.

\begin{table*}[h]
\footnotesize
    \centering
    \begin{tabular}{p{130pt}p{190pt}p{12pt}p{40pt}}
    \toprule
\textbf{Persona}&\textbf{Generated Stories}&\textbf{PC}&\textbf{Rouge-2}\\
\toprule
You have a loose grasp on your emotion and are quick to lash out. Often hurting the ones you care about most.&...\textbf{She had started to try to get her emotions under control, she had tried to keep herself calm during all of this, and she had only gotten worse.} She stood, just standing, and was staring at the class like an idiot, her head down...&0.90&0.0\\
\midrule
\underline{In command of} some measure of magical power&...Wilhelm is \underline{in command of} all forces in the north. After the war ended, his mind was full of food and drink, and he was ready for a quick trip to the bar. He had been to the bar on many occasions. Only in there that he could fully relax and forget about all those mess things...&0.16&0.14 \\
\bottomrule
\end{tabular}
\caption{Typical cases by PC~(Persona-Consistency) score. \textbf{Bold} words denote the consistent events adhering to the given persona. The overlapping words are \underline{underlined}.}
\label{tab:PC_cases}
\end{table*}

\section{Discussion of the Persona-Consistency Metric}~\label{appendix_pc}
To measure whether the generated story is consistent with the given persona, we propose the Persona-Consistency(PC) metric. In our experiments, we replace the ground-truth story with a randomly sampled one to construct a inconsistent story-persona pair as a negative sample. The fine-tuned classifier achieves an 83.63\% accuracy on the auto-constructed test set. However, it is possible that the PC metric depends on the word overlap to make predictions because of the simple random sampling of the negative samples~\cite{lin2020world}. We thus conduct a case study to investigate whether our PC metric depends on word overlap to make judgments. As shown in Table \ref{tab:PC_cases}, the first example gets a high PC score since the story embodies a consistent persona with the given persona description, in spite of a low rouge score. In contrast, the second example has an overlapped phrase ``in command of'' with the persona description but does not embody the corresponding persona description, and thus gets a high rouge score and low PC score. The results show that PC may not depend on shallow features like word overlap to make judgments.
    
What's more, we have taken into account the shortcomings of the automatic metrics for NLG and thus additionally added the human evaluation to further prove the effectiveness of our method. 

\end{document}